%% file: main.tex
\documentclass[10pt,twocolumn,letterpaper]{article}

\usepackage[pagenumbers]{cvpr} 

\input{preamble}

\definecolor{cvprblue}{rgb}{0.21,0.49,0.74}
\usepackage[pagebackref,breaklinks,colorlinks,allcolors=cvprblue]{hyperref}

\def\paperID{*****} 
\def\confName{CVPR}
\def\confYear{2026}

\title{MetroGS: Efficient and Stable Reconstruction of Geometrically Accurate High-Fidelity Large-Scale Scenes}

\author{
\textbf{Kehua Chen}$^{1,2}$,
\textbf{Tianlu Mao}$^{1,2}$,
\textbf{Xinzhu Ma}$^{3}$,
\textbf{Hao Jiang}$^{1,2\dag}$,
\textbf{Zehao Li}$^{1,2}$,
\textbf{Zihan Liu}$^{1,2}$,\\
\textbf{Shuqin Gao}$^{1}$,
\textbf{Honglong Zhao}$^{1}$,
\textbf{Feng Dai}$^{1}$,
\textbf{Yucheng Zhang}$^{1}$,
\textbf{Zhaoqi Wang}$^{1,2}$ \\
\\[-10pt]
$^1$Institute of Computing Technology, Chinese Academy of Sciences, ICT \\
$^2$University of Chinese Academy of Sciences, UCAS \;
$^3$Beihang University
}

\begin{document}

\twocolumn[{%
\maketitle
\begin{center}
\centering
\includegraphics[width=\textwidth]{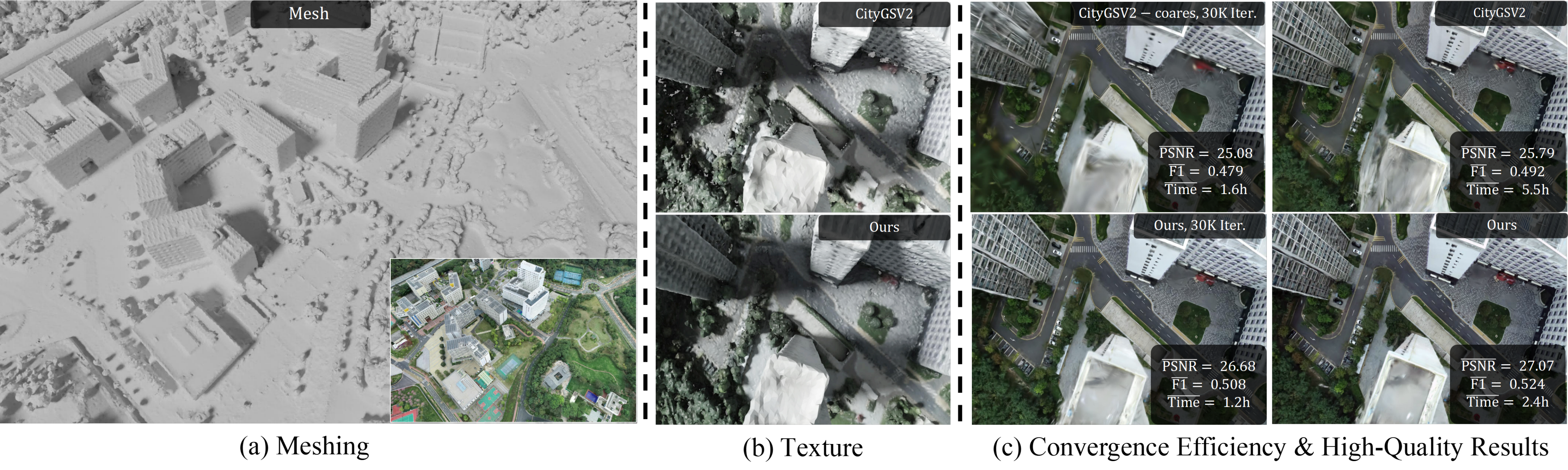}
\captionof{figure}{\textbf{Illustration of the superiority of our method.}
(a) Our method accurately reconstructs the geometric structure of large-scale urban scenes, faithfully restoring fine details such as buildings, vegetation, and roads.
(b) Compared with the SOTA method CityGSV2~\cite{liu2024citygaussianv2}, our result are more complete and geometrically precise.
(c) Benefiting from a well-designed training framework, our method achieves superior convergence speed and geometric quality.
On four RTX 3090 GPUs, our method reaches better performance with less than 25\% of the training time required by CityGSV2.
The quantitative results are reported based on the quality metrics of Modern Building~\cite{xiong2024gauu}.
}
\label{fig-top}
\end{center}
}]

\input{sec/0_abstract}    
\input{sec/1_intro}
\input{sec/2_formatting}
\input{sec/3_finalcopy}
{
    \small
    \bibliographystyle{ieeenat_fullname}
    \bibliography{main}
}
\input{sec/X_suppl}


\end{document}

%% file: preamble.tex
\usepackage{multirow}
\usepackage[table]{xcolor}
\usepackage{graphicx}

%% file: sec/0_abstract.tex
\begin{abstract}
Recently, 3D Gaussian Splatting and its derivatives have achieved significant breakthroughs in large-scale scene reconstruction. However, how to efficiently and stably achieve high-quality geometric fidelity remains a core challenge. To address this issue, we introduce MetroGS, a novel Gaussian Splatting framework for efficient and robust reconstruction in complex urban environments. Our method is built upon a distributed 2D Gaussian Splatting representation as the core foundation, serving as a unified backbone for subsequent modules.
To handle potential sparse regions in complex scenes, we propose a structured dense enhancement scheme that utilizes SfM priors and a pointmap model to achieve a denser initialization, while incorporating a sparsity compensation mechanism to improve reconstruction completeness.
Furthermore, we design a progressive hybrid geometric optimization strategy that organically integrates monocular and multi-view optimization to achieve efficient and accurate geometric refinement.
Finally, to address the appearance inconsistency commonly observed in large-scale scenes, we introduce a depth-guided appearance modeling approach that learns spatial features with 3D consistency, facilitating effective decoupling between geometry and appearance and further enhancing reconstruction stability.
Experiments on large-scale urban datasets demonstrate that MetroGS achieves superior geometric accuracy, rendering quality, offering a unified solution for high-fidelity large-scale scene reconstruction.
Project page: \href{https://m3phist0.github.io/MetroGS}{https://m3phist0.github.io/MetroGS}.
\end{abstract}

%% file: sec/1_intro.tex
\section{Introduction}
\label{sec:intro}

3D scene reconstruction is an essential topic in computer vision and graphics. Achieving large-scale, high-precision 3D modeling serves as the foundational support for numerous applications, such as aerial surveying~\cite{delplanque2024will, gu2023ue4}, autonomous driving~\cite{geiger2012we, caesar2020nuscenes, li2024gradiseg}, immersive AR/VR~\cite{wu20244d, chen2025dual, chen2025learning}. Recent achievements in 3D Gaussian Splatting (3DGS) have notably accelerated the translation of this goal toward practical application~\cite{kerbl20233d, yu2024mip, charatan2024pixelsplat, chen2024mixedgaussianavatar}, demonstrating remarkable rendering efficiency and visual fidelity. However, current methods, while excelling in rendering quality~\cite{li2025stdr, yuan2025robust, liu2025holistic, lin2024vastgaussian},
improvements in geometric reconstruction remain relatively limited, leading to an imbalance between visual fidelity and geometric accuracy. This imbalance highlights the need for a scalable reconstruction framework capable of preserving geometric accuracy under large-scale conditions.

In real-world urban environments, large-scale 3D reconstruction must cope with multiple challenges, including objects with diverse structures and scales, varying illumination conditions, and other complex factors.
Most existing methods primarily focus on scaling up 2DGS~\cite{huang20242d} or PGSR~\cite{chen2024pgsr} frameworks, yet their geometric optimization strategies remain underdeveloped.
Some methods~\cite{liu2024citygaussianv2, lin2025longsplat} rely solely on single-view constraints, making it difficult to maintain structural consistency. While others~\cite{chen2024gigags, li2025ulsr, chen20253d} adopt multi-view consistency constraints but typically use single-scale photometric constraints or simple reprojection errors, resulting in limited adaptability to complex large-scale environments.
Moreover, illumination and exposure inconsistencies are common in large-scale datasets~\cite{zhang2025ref, martin2021nerf}, forcing models to reconcile appearance variations during optimization, which compromises geometric consistency. Conventional multi-view consistency optimization struggles to effectively address such issues.
Meanwhile, we observe that insufficient initial sampling in weakly textured or sparsely observed regions is another key factor affecting geometric quality, often leading to inaccurate recovery of local structures and resulting in surface holes or structural artifacts.

To overcome these challenges, we propose MetroGS, a novel Gaussian Splatting framework that focuses on achieving \textbf{M}ulti-view \textbf{E}fficient \textbf{T}uning for \textbf{R}obust \textbf{O}ptimization in complex urban environments. 
Specifically, we adopt 2DGS as the core representation for modeling 3D geometry and utilize a distributed training strategy~\cite{zhao2024scaling} to efficiently support large-scale scene reconstruction. Building upon this foundation, we introduce a structured dense enhancement scheme. During initialization, the training images are partitioned based on SfM-derived priors, and a pre-trained pointmap model~\cite{wang2025pi} is employed to perform dense enhancement on the initial point cloud. In the subsequent densification stage, an additional sparse-compensation mechanism is incorporated to recover the remaining incomplete regions, thereby improving the overall reconstruction completeness and quality. Furthermore, we propose a progressive hybrid geometric refinement strategy.
During the early stage of training, we perform a lightweight monocular geometric optimization guided by priors from an off-the-shelf depth estimator~\cite{wang2025moge}. As training progresses, a multi-view refinement is introduced. Inspired by \cite{bleyer2011patchmatch, xu2019multi}, we adopt a carefully designed PatchMatch-based method to refine the rendered depths, and further complete the refined depth maps with monocular priors to obtain accurate and complete depth estimates for subsequent fine-grained geometric optimization.
This progressive design effectively balances geometric accuracy and computational efficiency. Finally, we address appearance inconsistencies by decoupling geometry and appearance. Specifically, we introduce a depth-guided appearance modeling module that adopts a Tri-Mip~\cite{hu2023tri} structure to store spatial features of the scene. By leveraging the high-quality optimized depth results, the module queries geometry-aligned 3D-consistent feature representations, thereby achieving efficient and stable appearance decoupling. Overall, these components form an efficient and consistent framework for large-scale scene reconstruction, and extensive experiments on multiple large-scale datasets validate its effectiveness.

Our main contributions can be summarized as follows:

\begin{itemize}
\item We design a structured dense enhancement scheme that optimizes initialization and densification to compensate for geometric deficiencies in sparse regions.
\item  We propose a progressive hybrid geometric refinement integrating monocular and PatchMatch-based multi-view optimization for efficient and accurate reconstruction.
\item We introduce a depth-guided appearance module that integrates geometry and appearance to mitigate inter-image variations and enhance reconstruction stability.
\item Comprehensive experiments show that our method delivers superior reconstruction quality across diverse large-scale scenes.
\end{itemize}

\section{Related Works}

\subsection{Novel View Synthesis}

Novel view synthesis aims to generate high-fidelity images from arbitrary viewpoints by learning an underlying three-dimensional scene representation. The pivotal work NeRF~\cite{mildenhall2021nerf} implicitly models the scene using MLPs to encode color and density information for 3D points and viewing directions, enabling novel view synthesis of complex scenes. To address performance limitations, methods exemplified by Tri-MipRF~\cite{hu2023tri} employed more advanced feature encoding techniques~\cite{muller2022instant, chen2022tensorf} to improve both efficiency and rendering quality. More recently, 3DGS~\cite{kerbl20233d} emerged as another influential framework, which models scenes using explicit 3D Gaussian primitives and achieves real-time rendering through differentiable rasterization. Following this breakthrough, a series of works~\cite{zhang2024pixel, fang2024mini, ye2024absgs} primarily focus on enhancing rendering quality. These advances collectively inspire the design of our proposed algorithm.

\subsection{Surface Reconstruction}
The ability of surface reconstruction to generate accurate 3D geometry from diverse inputs is critical for realizing practical 3D technology. Recently, many advanced methods~\cite{guedon2024sugar, yu2024gaussian, wang2024gaussurf} have been developed extended from 3DGS. PGSR~\cite{chen2024pgsr} achieves high-fidelity and efficient surface reconstruction by introducing a planar Gaussian representation combined with unbiased depth rendering and multi-view geometric regularization. 2DGS~\cite{huang20242d} enhances geometric accuracy by substituting 3D Gaussians with surface-oriented 2D surfels, addressing multi-view inconsistency inherent in 3DGS, and serves as the foundational approach adopted by the best current surface reconstruction methods~\cite{huang2025fatesgs, wu2025sparse2dgs, zhang2025ref}. 
Nevertheless, these approaches are mainly optimized for object-level scenes and cannot be directly applied to large-scale scenes with reliable performance.

\subsection{Large Scale Scene Reconstruction}
The task of large-scale reconstruction demands coping with vast amounts of data and more complex scene environments. Several recent works~\cite{ren2024octree, lin2024vastgaussian, zhao2024scaling, liu2025holistic, yuan2025robust} have extended 3DGS to large-scale scenes, focusing on rendering quality and efficiency improvements. In contrast, the exploration dedicated to surface reconstruction remains at a relatively early stage. CityGSV2~\cite{liu2024citygaussianv2} continued the strategy of partitioned parallel training, optimizing 2DGS to adapt it for large-scale scene reconstruction, and simultaneously established standard geometric benchmarks for large-scale scenes. CityGS-$\mathcal{X}$~\cite{gao2025citygs} introduced a scalable architecture supporting multi-GPU parallel rendering, and jointly optimizes the scene's geometry and appearance through batch-level multi-task training. While other methods~\cite{gao2024cosurfgs, chen2024gigags} have extended surface reconstruction algorithms to large-scale scenes, their simple geometric optimization struggles with stability in complex large scenes. Our method advances this field by introducing targeted geometric optimization for more robust and higher-quality outcomes.

\section{Preliminaries}
3D Gaussian Splatting~\cite{kerbl20233d} models a scene as anisotropic Gaussian primitives, each defined by its center, covariance, opacity, and SH coefficients for view-dependent color. Rendering is performed via front-to-back $\alpha$-blending of the $\alpha$-weighted contributions along each ray:
\begin{equation}
    C = \sum_{i \in N} c_i \alpha_i T_i, \quad T_i = \prod_{j=1}^{i-1} (1-\alpha_j).
\end{equation}
2D Gaussian Splatting~\cite{huang20242d} flattens the 3D ellipsoid volume into 2D planar disks, making the primitives highly suitable for explicit 3D surface representation and optimization. In 2DGS, depth calculation is primarily divided into mean depth and median depth. The latter is considered more robust, as it utilizes visibility and treats $T_i=0.5$ as the pivot for surface and free space:
\begin{equation}
    D = max\{z_i|T_i>0.5\}.
\end{equation}
A regularization term aligning depth gradients with normals enables 2DGS to achieve geometric optimization.

%% file: sec/2_formatting.tex
\begin{figure*}[!htbp]
\centering
\includegraphics[width=1\textwidth]{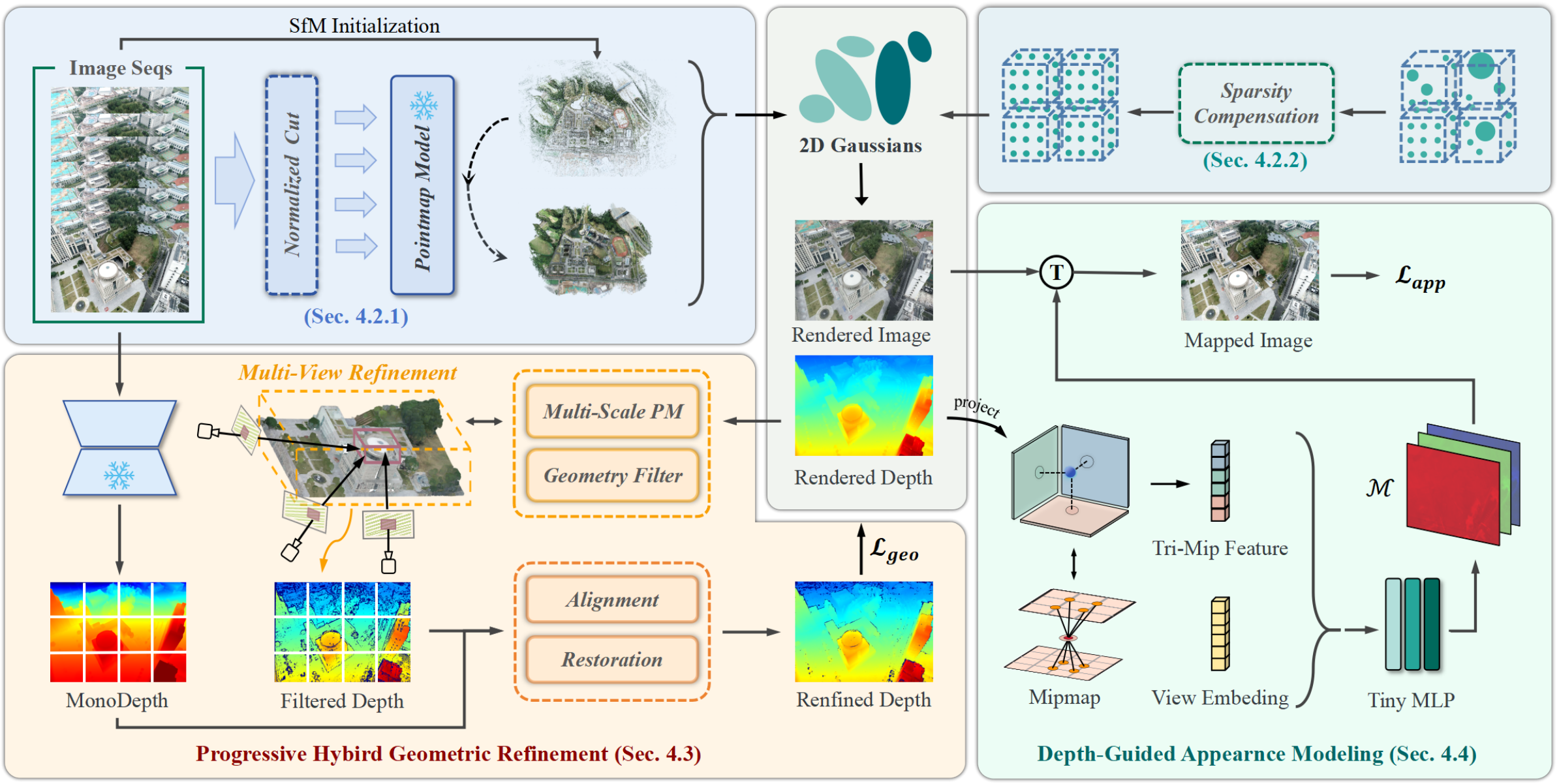}
\caption{{\bf Overview.} Starting with the input image sequences, we first utilize the prior information provided by SfM, combined with a pointmap model, to generate a high-quality initial point cloud. Next, an additional sparsity compensation optimization is introduced during the densification process to further refine sparse regions. We then combine monocular depth priors with multi-view consistency optimization to achieve progressive hybrid geometric refinement. Simultaneously, a depth-guided appearance modeling module is employed to decouple geometry and appearance, thereby enhancing reconstruction fidelity.}
\label{overview}
\end{figure*}

\section{Method}
\label{sec:method}
Large-scale surface reconstruction tasks face multiple challenges, including the vast spatial extent of scenes, the insufficient quality of initial reconstruction points, the structural diversity and complexity of objects, and the heterogeneity of image data caused by inconsistent lighting conditions. 

To address these challenges, we propose an efficient and highly robust framework for large-scale scene reconstruction. An overview of our method is shown in Fig.~\ref{overview}. The main components of our method are structured as follows: Section~\ref{Para} first introduces our fundamental parallel training framework. Following this, the subsequent sections detail the key mechanisms designed to achieve high-precision reconstruction. Specifically, Section~\ref{Dense} elaborates on our structured dense enhancement scheme. Section~\ref{Geo} then describes the progressive hybird geometric refinement method. Finally, Section~\ref{App} presents the depth-guided appearance modeling.
\subsection{Scalable Parallel Strategy} \label{Para}
We extend 2DGS into a Gaussian-wise distributed training paradigm inspired by the parallel concepts in~\cite{zhao2024scaling}. Specifically, the initialization point cloud is uniformly distributed across multiple GPUs for local Gaussian initialization, and multi-view batched training is employed to evenly assign images among devices. Each worker leverages the spatial locality of Gaussian Splatting to fetch only the required Gaussian subsets, enabling efficient communication. During dynamic densification, load balance is maintained through periodic Gaussian redistribution. This distributed design maximizes computational resource utilization and demonstrates excellent scalability, allowing efficient support for large-scale scene reconstruction.

\subsection{Structured Dense Enhancement} \label{Dense}
Gaussian initialization is based on 3D points from SfM~\cite{schonberger2016structure}. However, even in large-scale scenes with dense coverage, the presence of sparse-view or weak-texture regions leads to an overly sparse initial point cloud. To mitigate this, we introduce a structured dense enhancement scheme that separately optimizes initialization and densification.

\subsubsection{Pointmap Model Assisted Initialization} 
We incorporate the pointmap model~\cite{wang2025pi} to obtain auxiliary initial dense point clouds for Gaussian initialization, leveraging its capability for efficient 3D structure prediction. We first construct an undirected image graph $G=(V,E)$, where each node represents an image and each edge weight $w_{ij}$ corresponds to the number of inter-image feature matches estimated by SfM. The graph is partitioned into $N$ clusters, matching the number of available GPUs, by minimizing the normalized cut objective:
\begin{equation}
\mathrm{Ncut}(A_1, \ldots, A_N)
= \sum_{k=1}^{N}
\frac{
\mathrm{Cut}(A_k, \bar{A}_k)
}{
\mathrm{Vol}(A_k)
},
\label{eq:ncut}
\end{equation}
where $\mathrm{Cut}(A_k,\bar{A}_k)$ and $\mathrm{Vol}(A_k)$ represent the inter-cluster and intra-cluster connection weights, respectively. This criterion encourages clusters with strong intra-cluster connectivity and weak inter-cluster links. Subsequently, we apply the pointmap model to these clusters in parallel for dense 3D prediction. Within each cluster, images are ordered according to their matching connectivity and processed in mini-batches. After each batch, pixel indices provide one-to-one 3D correspondences between the dense pointmap and the SfM reconstruction. We then estimate a similarity transformation matrix $\mathbf T^{\ast}$ to align the dense prediction with the SfM coordinate frame:
\begin{equation}
\mathbf T^{\ast}
=\arg\min_{\mathbf T\in\mathrm{Sim}(3)}
\left\|\, \mathbf T\,\tilde{\mathbf X}-\tilde{\mathbf Y} \right\|_{F}^{2},
\end{equation}
where $\tilde{\mathbf X},\tilde{\mathbf Y}\in\mathbb R^{4\times m}$ are the homogeneous representations of the dense and SfM 3D points. Finally, all aligned cluster results are sampled and merged into a unified auxiliary point cloud  for Gaussian initialization.

\subsubsection{Sparsity Compensation Densification}

When the initialized regions are excessively sparse, they tend to form large, coarse Gaussian primitives. If such regions are observed by only a few effective views, the resulting representations are difficult to densify properly. To address this issue, we introduce a targeted optimization strategy designed to refine and densify these under-represented areas. We identify Gaussians $\mathbf{G}_{\text{split}}$ for splitting based on a dual criterion combining large contribution area and low local density:
\begin{equation}
\mathbf{G}_{\text{split}}
=\left\{\, G_i \;\middle|\; S_i > S_{\text{th}} \;\land\; V_i < V_{\text{th}} \,\right\}.
\end{equation}
Here, $S_i = \sum_{x\in\mathcal P} \delta\!\bigl(i_{\max}(x)=i \land i_{\text{mid}}(x)=i\bigr)$ denotes the accumulated area where Gaussian $G_i$ simultaneously yields the maximum contribution weight and the median depth along the ray.
$V_i$ measures the local voxel density, defined as the number of Gaussians whose centers fall within the voxel $\mathbf V_{G_i}$ containing $G_i$.
This criterion favors splitting Gaussians that dominate large regions yet lie in sparse neighborhoods, thereby improving geometric coverage without over-densification.

\subsection{Progressive Hybrid Geometric Refinement} \label{Geo}

Robust geometric optimization is key to high-quality surface reconstruction. Traditional methods rely on monocular depth supervision or multi-view photometric constraints. However, the former lacks inter-view geometric consistency, while the latter, being single-scale and computationally demanding, is limited in structurally diverse scenes. To address this, we propose a two-stage progressive hybrid geometric refinement strategy.

\subsubsection{Single-View Optimization}
Following~\cite{liu2024citygaussianv2}, we employ a pretrained depth estimation model~\cite{wang2025moge} to obtain a monocular depth prior. The estimated inverse depth is first aligned with the sparse SfM depth, and the L1 loss between the rendered and estimated inverse depths is formulated as $\mathcal{L}_d$ to guide depth supervision. In addition, we preserve the depth–normal consistency loss $\mathcal{L}_n$ from 2DGS~\cite{huang20242d} to further enhance geometric fidelity.
In practice, we also observe that large-scale Gaussians often introduce noticeable visual artifacts and blur local details, and their extensive coverage on the image plane leads to heavy GPU memory consumption during training and slows down the optimization process. To mitigate these issues, we introduce a scale regularization term defined as: 
\begin{equation}
    \mathcal{L}_s = \frac{1}{|M|}\sum_{i\in M} max(max(s_i)-\tau_s, \epsilon),
\end{equation}
where $M$ denotes the set of visible Gaussians, and $\tau_s$ is a threshold that limits the maximum allowable Gaussian scale. The overall geometry optimization loss at this stage is formulated as:
\begin{equation}
    \mathcal{L}^{(1)}_{geo} = \lambda_d\mathcal{L}_d + \lambda_n\mathcal{L}_n + \lambda_s\mathcal{L}_s. 
\end{equation}

\subsubsection{Hybird Multi-View Refinement}
After sufficient training iterations, the geometric optimization transitions into the second stage. For each training image, we predefine its neighboring views based on the prior information provided by SfM. For each training image, we refine its rendered depth $\mathcal{D}_r$ using the PatchMatch algorithm between the image and its neighboring views. To effectively handle objects of different scales, we iteratively apply multi-scale patches for depth refinement.
The refined depth is further filtered based on reprojection errors with adjacent views, yielding the final reliable depth $\mathcal{D}_f$. 

\begin{figure}
    \centering
    \includegraphics[width=1.\linewidth]{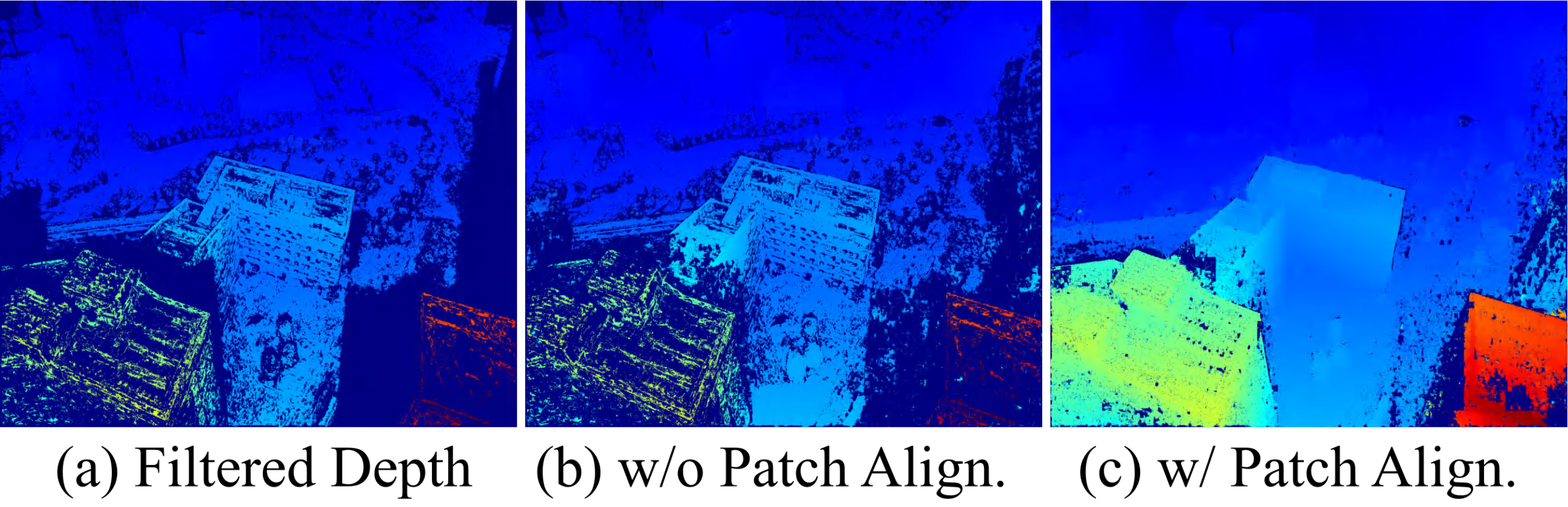}
    \caption{\textbf{Visualization of hybrid multi-view refinement.} (a) Strict geometric consistency yields reliable PM-refined depth. (b) and (c) show the restored refined depths, highlighting the effectiveness of patch-based alignment for local restoration.}
    \label{fig:fig-align}
\end{figure}

Although this filtering effectively mitigates the impact of incorrect depths, it inevitably removes some valid regions, resulting in large holes in the refined depth maps.
To alleviate this issue, we reintroduce the monocular depth prior $\mathcal{D}_m$, leveraging its relatively accurate relative depth estimation as complementary guidance to recover valid depth regions that were mistakenly filtered out. Specifically, the monocular depth map is divided into local patches, and each patch is locally aligned with its corresponding filtered depth via least-squares estimation:
\begin{equation}\label{eq8}
s^{\ast},\,t^{\ast}
=\arg\min_{s,\,t}
\sum_{p\in \mathcal{D}_{f}}\|\mathcal{D}_{f}(p) - (s\,\cdot \mathcal{D}_{m}(p)+t)\|^{2}.
\end{equation}
When the alignment error between the aligned depth and the filtered depth falls below a predefined threshold, the filtered depth is preserved. The restored depth $\mathcal{D}_{mv}$ is then used to guide further refinement of the rendered depth $\mathcal{D}_r$. Specifically, the depth refinement loss is defined as:
\begin{equation}
    \mathcal{L}_{mv} = \frac{1}{|\mathcal{D}_{mv}|} \sum_{p \in \mathcal{D}_{mv}} \left| \mathcal{D}_{r}(p) - \mathcal{D}_{mv}(p) \right|.
\end{equation}
Unlike direct photometric optimization, we adopt depth-based supervision for enforcing multi-view consistency. This design provides two benefits. As the quality of the rendered depth improves during training, the refined depth improves accordingly, and computation is reduced by updating the refined depth maps only at fixed intervals. These combined mechanisms ensure that the final depth maps achieve both high geometric accuracy and structural completeness. The overall geometry optimization loss at this stage is formulated as:
\begin{equation}
\mathcal{L}^{(2)}_{geo}=\lambda_{mv}\mathcal{L}_{mv}+\lambda_n\mathcal{L}_n.
\end{equation}

\begin{table*}[htp]
  \caption{\textbf{Comparison with SOTA reconstruction methods on the GauU-Scene~\cite{xiong2024gauu} dataset.} P and R indicate the Precision and Recall with respect to the ground-truth point cloud. Results highlighted in \colorbox{red!25}{red}, \colorbox{orange!30}{orange}, and \colorbox{yellow!40}{yellow} correspond to the best, second-best, and third-best performances, respectively. ``NaN" means no results due to NaN error. ``FAIL" means fail to extract meaningful mesh.}
  \label{tab:gauu}
  \centering
  \setlength{\tabcolsep}{0.9pt}
  \renewcommand{\arraystretch}{1.1}
  \resizebox{\textwidth}{!}{
  \begin{tabular}{l|
>{\centering\arraybackslash}p{1cm}
>{\centering\arraybackslash}p{1cm}
>{\centering\arraybackslash}p{1.1cm}
|>{\centering\arraybackslash}p{1cm}
>{\centering\arraybackslash}p{1cm}
>{\centering\arraybackslash}p{1cm}
|>{\centering\arraybackslash}p{1cm}
>{\centering\arraybackslash}p{1cm}
>{\centering\arraybackslash}p{1.1cm}
|>{\centering\arraybackslash}p{1cm}
>{\centering\arraybackslash}p{1cm}
>{\centering\arraybackslash}p{1cm}
|>{\centering\arraybackslash}p{1cm}
>{\centering\arraybackslash}p{1cm}
>{\centering\arraybackslash}p{1.1cm}
|>{\centering\arraybackslash}p{1cm}
>{\centering\arraybackslash}p{1cm}
>{\centering\arraybackslash}p{1cm}}
    \toprule
    \multirow{2}{*}{Methods} 
    & \multicolumn{6}{c}{Russian Building} 
    & \multicolumn{6}{|c}{Residence} 
    & \multicolumn{6}{|c}{Modern Building} \\
    \cmidrule(lr){2-7} \cmidrule(lr){8-13} \cmidrule(lr){14-19}
     & PSNR$\uparrow$ & SSIM$\uparrow$ & LPIPS$\downarrow$  & P$\uparrow$ & R$\uparrow$ & F1$\uparrow$ 
     & PSNR$\uparrow$ & SSIM$\uparrow$ & LPIPS$\downarrow$  & P$\uparrow$ & R$\uparrow$ & F1$\uparrow$ 
     & PSNR$\uparrow$ & SSIM$\uparrow$ & LPIPS$\downarrow$  & P$\uparrow$ & R$\uparrow$ & F1$\uparrow$ \\
    \midrule
    NeuS
    & 13.65 & 0.202 & 0.694 & FAIL & FAIL & FAIL 
    & 15.16 & 0.244 & 0.674 & FAIL & FAIL & FAIL 
    & 14.58 & 0.236 & 0.694 & FAIL & FAIL & FAIL \\
    Neuralangelo
    & 12.48 & 0.328 & 0.698 & NaN & NaN & NaN
    & NaN & NaN & NaN & NaN & NaN & NaN 
    & NaN & NaN & NaN & NaN & NaN & NaN \\ 
    SuGaR
    & 23.62 & 0.738 & 0.332 & 0.480 & 0.369 & 0.417 
    & 21.95 & 0.612 & 0.452 & \cellcolor{red!25}0.579 & 0.287 & 0.384
    & 24.92 & 0.700 & 0.381 & \cellcolor{orange!30}0.650 & 0.220 & 0.329 \\
    GOF
    & 21.30 & 0.713 & 0.322 & 0.294 & 0.394 & 0.330 
    & 20.68 & 0.652 & 0.391 & 0.404 & \cellcolor{yellow!40}0.418 & 0.411 
    & 25.01 & 0.749 & 0.286 & 0.411 & 0.357 & 0.382 \\
    2DGS
    & 23.77 & 0.788 & 0.189 & 0.544 & \cellcolor{yellow!40}0.519 & \cellcolor{yellow!40}0.531 
    & 22.24 & 0.703 & 0.306 & 0.526 & 0.406 & 0.458 
    & 25.77 & 0.776 & 0.202 & 0.588 & \cellcolor{orange!30}0.413 & \cellcolor{yellow!40}0.485 \\
    CityGS
    & \cellcolor{yellow!40}24.37 & \cellcolor{orange!30}{0.808} & \cellcolor{yellow!40}0.163 & 0.459 & 0.443 & 0.451 
    & \cellcolor{yellow!40}23.59 & \cellcolor{orange!30}{0.763} & \cellcolor{orange!30}{0.204} & 0.524 & 0.391 & 0.448 
    & \cellcolor{orange!30}{26.29} & \cellcolor{orange!30}{0.796} & \cellcolor{orange!30}{0.160} & 0.582 & 0.381 & 0.461 \\
    CityGS-$\mathcal{X}$ 
    & \cellcolor{orange!30}{24.62} & \cellcolor{yellow!40}0.804 & \cellcolor{orange!30}{0.155} & \cellcolor{orange!30}0.570 & 0.497 & \cellcolor{yellow!40}0.531 
    & \cellcolor{orange!30}{23.74} & \cellcolor{yellow!40}0.749 & \cellcolor{yellow!40}0.238 & \cellcolor{yellow!40}0.564 & 0.402 & \cellcolor{orange!30}0.470 
    & \cellcolor{yellow!40}26.20 & \cellcolor{yellow!40}0.783 & \cellcolor{yellow!40}0.171 & 0.598 & 0.362 & 0.451 \\
    CityGSV2 
    & 24.12 & 0.784 & 0.196 & \cellcolor{yellow!40}0.560 & \cellcolor{orange!30}0.530 & \cellcolor{orange!30}0.544 
    & 23.57 & 0.742 & 0.243 & 0.524 & \cellcolor{orange!30}0.421 & \cellcolor{yellow!40}0.467 
    & 25.84 & 0.770 & 0.207 & \cellcolor{yellow!40}0.643 & \cellcolor{yellow!40}0.398 & \cellcolor{orange!30}0.492 \\
    Ours 
    & \cellcolor{red!25}24.94 & \cellcolor{red!25}0.814 & \cellcolor{red!25}0.138 & \cellcolor{red!25}0.610 & \cellcolor{red!25}0.562 & \cellcolor{red!25}0.585 
    & \cellcolor{red!25}24.51 & \cellcolor{red!25}0.769 & \cellcolor{red!25}0.185 & \cellcolor{orange!30}0.566 & \cellcolor{red!25}0.439 & \cellcolor{red!25}0.494 
    & \cellcolor{red!25}27.07 & \cellcolor{red!25}0.797 & \cellcolor{red!25}0.152 & \cellcolor{red!25}0.662 & \cellcolor{red!25}0.433 & \cellcolor{red!25}0.524 \\
    \bottomrule
  \end{tabular}
} 
\end{table*}

\begin{figure}[t]
\centering
\includegraphics[width=1\linewidth]{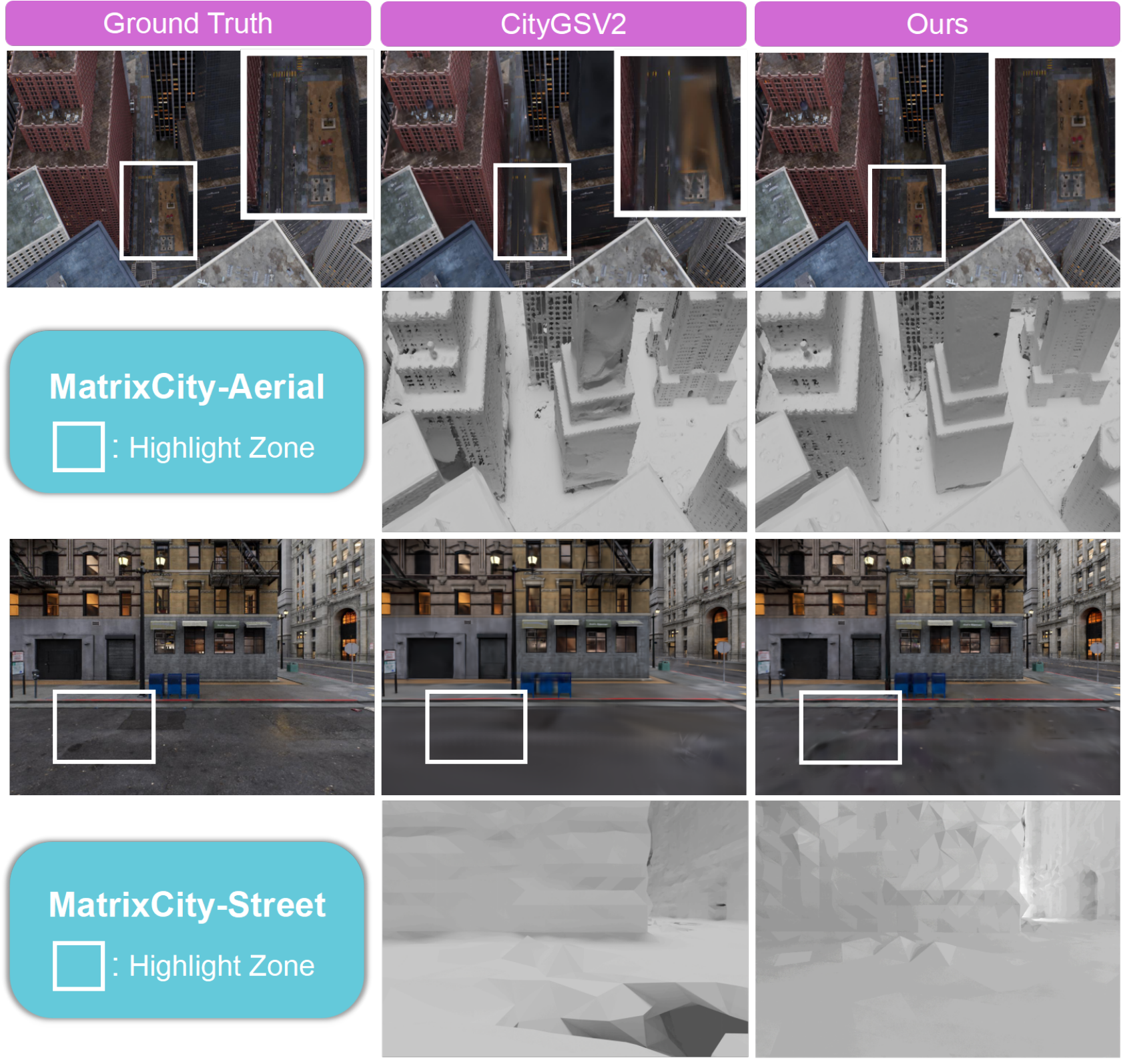} 
\caption{\textbf{Qualitative comparison on the MatrixCity~\cite{li2023matrixcity} dataset.} Image rendering and mesh reconstruction are compared between our method and CityGSV2~\cite{liu2024citygaussianv2}.}
\label{fig2}
\end{figure}

\subsection{Depth-Guided Appearance Modeling} \label{App}

Previous works~\cite{lin2024vastgaussian, zhang2025ref} show that accurate appearance modeling is crucial for realistic reconstruction, since geometry-only methods often struggle under complex imaging conditions. Existing appearance methods typically do not leverage geometric information, while our method provides high-quality rendered depth that offer a reliable structural prior for appearance learning. Based on this, we design a depth-guided appearance modeling module to ensure appearance estimation under precise geometric constraints, enabling true geometry–appearance decoupling.

Specifically, we employ a Tri-Mip~\cite{hu2023tri} structure to store scale-adaptive, multi-resolution 3D features of the scene, which maintain cross-view consistency in space. Given the rendered depth map $D_r$, we query the Tri-Mip feature planes using the 3D coordinates of each pixel's projection, resulting in structure-aligned representations $f_{\text{Tri}}(x)$ . These features provide a stable geometric foundation for appearance estimation, enabling it to focus on color and lighting variations that are independent of geometry. Additionally, each training image $\mathcal{I}_i$ is assigned a learnable appearance embedding $l_i \in \mathbb{R}^d$ to capture global illumination and exposure conditions. 
The queried Tri-Mip feature $f_{\text{Tri}}(x)$ and the embedding $l_i$ are concatenated and passed through a lightweight MLP tone mapper
$\mathcal{F}_{\theta}$:
\begin{equation}
\mathcal{M}(x) = \mathcal{F}_{\theta}\big(f_{\text{Tri}}(x);\, l_i\big),
\end{equation}
where $\mathcal{M}(x)$ denotes the tone-mapped appearance at pixel $x$. 
The output is used to modulate the rendered image $\mathcal{I}^r_i$, resulting in the final reconstruction $\mathcal{I}^t_i$ with consistent tone and illumination. The appearance loss is defined as:
\begin{equation}
    \mathcal{L}_{app} = \lambda \mathcal{L}_1(\mathcal{I}_i^t, {\mathcal{I}_i}) + (1-\lambda)\mathcal{L}_{D-SSIM}(\mathcal{I}_i^r, \mathcal{I}_i).
\end{equation}

\begin{table}[t]
  \footnotesize
  \caption{\textbf{Comparison on the MatrixCity ~\cite{li2023matrixcity} dataset.} “–” indicates the metric was not reported in the original paper.}
  \label{tab:matrixcity}
  \centering
  \setlength{\tabcolsep}{2pt}
  \renewcommand{\arraystretch}{.96}
  \resizebox{\linewidth}{!}{
  \begin{tabular}{l|
>{\centering\arraybackslash}p{1cm}
>{\centering\arraybackslash}p{1cm}
>{\centering\arraybackslash}p{1.1cm}
|>{\centering\arraybackslash}p{1cm}
>{\centering\arraybackslash}p{1cm}
>{\centering\arraybackslash}p{1cm}}
    \toprule
    \multirow{2}{*}{Methods} 
    & \multicolumn{6}{c}{MatrixCity-Aerial} \\
    \cmidrule(lr){2-7}
     & PSNR$\uparrow$ & SSIM$\uparrow$ & LPIPS$\downarrow$ & P$\uparrow$ & R$\uparrow$ & F1$\uparrow$ \\
    \midrule
    SuGaR
    & 22.41 & 0.633 & 0.493 & 0.182 & 0.157 & 0.169 \\
    GOF
    & 17.42 & 0.374 & 0.588 & FAIL & FAIL & FAIL \\
    2DGS
    & 21.35 & 0.632 & 0.562 & 0.207 & 0.390 & 0.270 \\
    CityGS
    & \cellcolor{yellow!40}{27.46} & \cellcolor{red!25}{0.865} & \cellcolor{yellow!40}0.204 & 0.362 & 0.637 & 0.462 \\
    CityGS-$\mathcal{X}$ 
    & \cellcolor{red!25}27.58 & $-$ & $-$ & \cellcolor{orange!30}0.444 & \cellcolor{red!25}0.840 & \cellcolor{orange!30}0.581 \\
    CityGSV2 
    & 27.23 & \cellcolor{orange!30}0.857 & \cellcolor{orange!30}{0.169} & \cellcolor{yellow!40}0.441 & \cellcolor{yellow!40}0.752 & \cellcolor{yellow!40}0.556 \\
    Ours 
    & \cellcolor{orange!30}27.52 & \cellcolor{yellow!40}0.854 & \cellcolor{red!25}{0.167} & \cellcolor{red!25}0.572 & \cellcolor{orange!30}0.828 & \cellcolor{red!25}0.677 \\
    \midrule
    \multirow{2}{*}{Methods} 
    & \multicolumn{6}{c}{MatrixCity-Street} \\
    \cmidrule(lr){2-7}
     & PSNR$\uparrow$ & SSIM$\uparrow$ & LPIPS$\downarrow$ & P$\uparrow$ & R$\uparrow$ & F1$\uparrow$ \\
    \midrule
    SuGaR
    & 19.82 & 0.662 & 0.478 & 0.053 & 0.111 & 0.071 \\
    GOF
    & 20.32 & 0.703 & 0.440 & 0.219 & 0.473 & 0.300 \\
    2DGS
    & 21.50 & 0.723 & 0.477 & \cellcolor{yellow!40}0.334 & 0.659 & \cellcolor{yellow!40}0.443 \\
    CityGS
    & \cellcolor{orange!30}{22.98} & \cellcolor{red!25}{0.808} & \cellcolor{orange!30}0.301 & 0.283 & \cellcolor{yellow!40}0.689 & 0.401 \\
    CityGS-$\mathcal{X}$ 
    & OOM & OOM & OOM & OOM & OOM & OOM \\
    CityGSV2 
    & \cellcolor{yellow!40}22.24 & \cellcolor{yellow!40}0.788 & \cellcolor{yellow!40}{0.347} & \cellcolor{orange!30}0.376 & \cellcolor{orange!30}0.759 & \cellcolor{orange!30}0.503 \\
    Ours 
    & \cellcolor{red!25}23.16 & \cellcolor{orange!30}0.798 & \cellcolor{red!25}{0.294} & \cellcolor{red!25}0.480 & \cellcolor{red!25}0.828 & \cellcolor{red!25}0.607 \\
    \bottomrule
  \end{tabular}
} 
\end{table}

\subsection{Training Loss}

During training, the geometric and appearance optimization processes are jointly performed, and the geometric loss varies with the training stage. The overall loss function is defined as follows:
\begin{equation}
    \mathcal{L}_{total} = \mathcal{L}_{geo} + \mathcal{L}_{app}.
\end{equation}

\begin{figure*}[t]
\centering
\includegraphics[width=.99\textwidth]{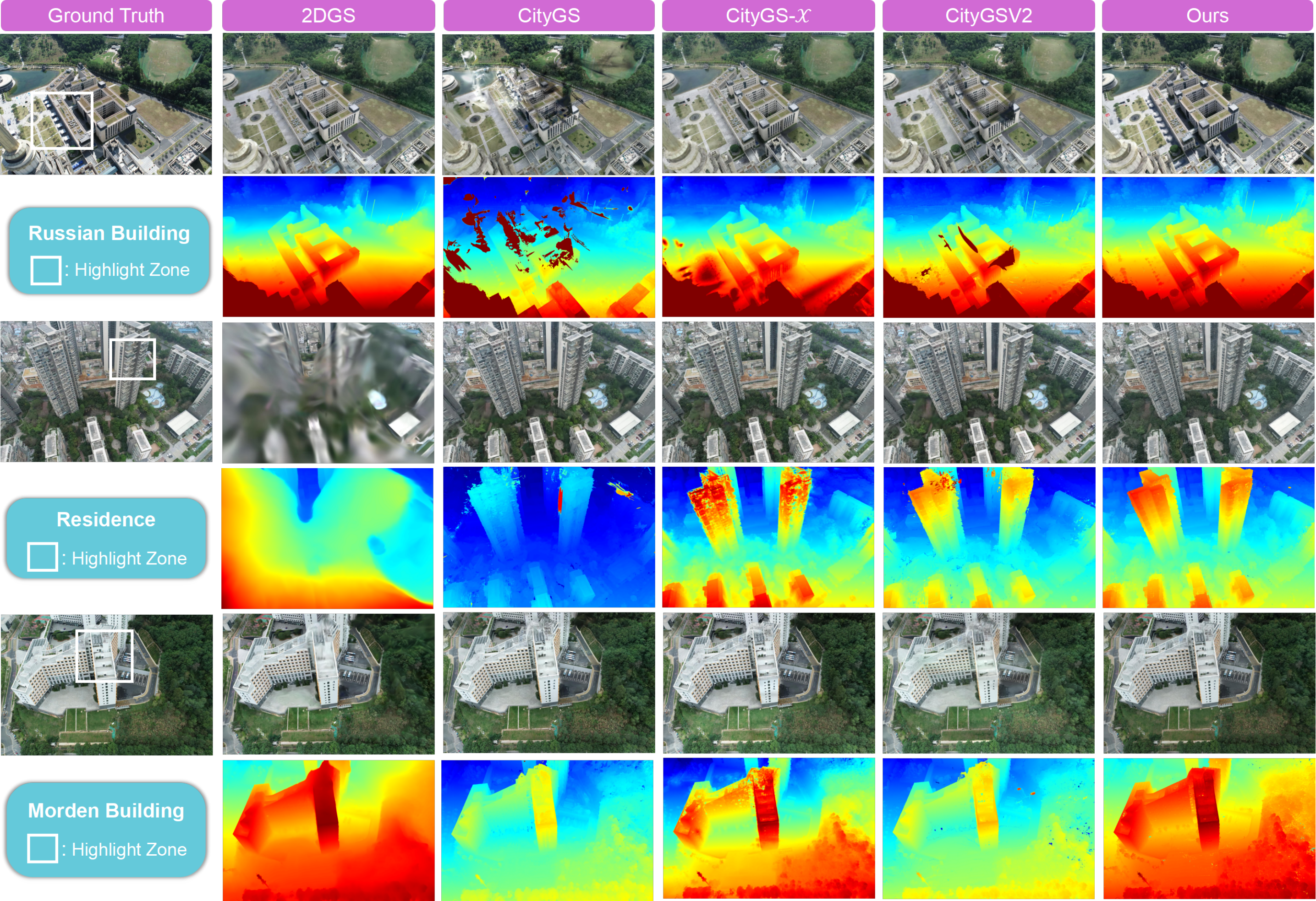} 
\caption{\textbf{Qualitative results on the GauU-Scene~\cite{xiong2024gauu} dataset.} We present the image and depth rendering results of our method compared with state-of-the-art methods.}
\label{fig1}
\end{figure*}

\section{Experiments}

\subsection{Experimental Setup}

We conduct comprehensive experiments on the GauU-Scene~\cite{xiong2024gauu} dataset and the synthetic MatrixCity~\cite{li2023matrixcity} dataset. All experiments are performed on a workstation equipped with four RTX 3090 GPUs. Both datasets provide high-precision ground-truth point clouds, making them reliable benchmarks for evaluating geometric quality in large-scale scene reconstruction.
Following the settings in~\cite{liu2024citygaussianv2}, we evaluate our method on both the aerial-view and street-view versions. The aerial-view images are downsampled to a 1600-pixel long side, while the street-view images retain their original $1000 \times 1000$ resolution.  
For mesh extraction, we adopt the 2DGS methodology, integrating median depth with TSDF fusion. Detailed training and evaluation configurations are provided in the supplementary materials.

\subsection{Baselines}

We compare our method against a broad set of state-of-the-art surface reconstruction methods. For NeRF-based methods, we include NeuS~\cite{wang2021neus} and Neuralangelo~\cite{li2023neuralangelo}. For GS–based methods, we compare against SuGaR~\cite{guedon2024sugar}, 2DGS~\cite{huang20242d}, GOF~\cite{yu2024gaussian}, CityGS~\cite{liu2024citygaussian}, CityGS-$\mathcal{X}$~\cite{gao2025citygs}, and CityGSV2~\cite{liu2024citygaussianv2}. We assess reconstruction quality from both visual and geometric perspectives. For large-scale scene reconstruction, we select CityGSV2 as a representative baseline, as it is among the best-performing open-source methods in terms of geometric reconstruction quality.

\subsection{Main Results}

\paragraph{Quantitative Results.}
As shown in Tab.~\ref{tab:gauu}, we compare our proposed method with several SOTA methods on the GauU-Scene~\cite{xiong2024gauu} dataset, which contains representative real-world urban scenes. The results show that our method achieves superior geometric reconstruction and rendering performance, ranking first on most metrics. Compared to CityGSV2~\cite{liu2024citygaussianv2}, our method improves PSNR by 0.88 dB on average and boosts the F1-score by 0.033, consistently outperforming it across all metrics. Tab.~\ref{tab:matrixcity} presents the comparison results on the synthetic MatrixCity~\cite{li2023matrixcity} dataset. Our method again achieves the highest F1-score, with an average improvement of 0.11 over CityGSV2, indicating reliable and accurate geometric reconstruction across different scene types and data settings. In addition, even under the inherently stable illumination of synthetic data, our method maintains competitive rendering performance. Overall, these results validate the effectiveness of our method for robust, geometrically accurate, and high-fidelity large-scale scene reconstruction. 


\paragraph{Qualitative Results.} 
To further validate the effectiveness of our method, we provide extensive visual comparisons. 
Fig.~\ref{fig2} shows image renderings and mesh reconstructions from synthetic MatrixCity~\cite{li2023matrixcity}, where our method produces more accurate and more complete geometric reconstructions than CityGSV2~\cite{liu2024citygaussianv2}. 
Fig.~\ref{fig1} presents qualitative comparisons of rendered images and corresponding depth maps from GauU-Scene~\cite{xiong2024gauu}. 
The first row depicts a scene under challenging lighting conditions: other methods suffer from floating artifacts caused by geometric errors, whereas our method yields more accurate illumination and more consistent geometric structures. 
The second and third rows contain scenes with rich texture details, where other methods show blurred or distorted structures in both the rendered images and depth maps. 
In contrast, our method preserves fine geometry and delivers visually coherent rendering results.

\subsection{Ablation Studies}
We conduct thorough ablation studies on the Russian Building scene to quantify the effectiveness of our proposed components, with results presented in Tab.~\ref{tab:ablation}.

\vspace{-2mm}
\paragraph{Scalable Parallel Strategy.} The first two rows show that the parallelization strategy substantially improves overall performance and running efficiency, which is critical for the high-performance execution of our framework.

\begin{figure}[t]
\centering
\includegraphics[width=1\linewidth]{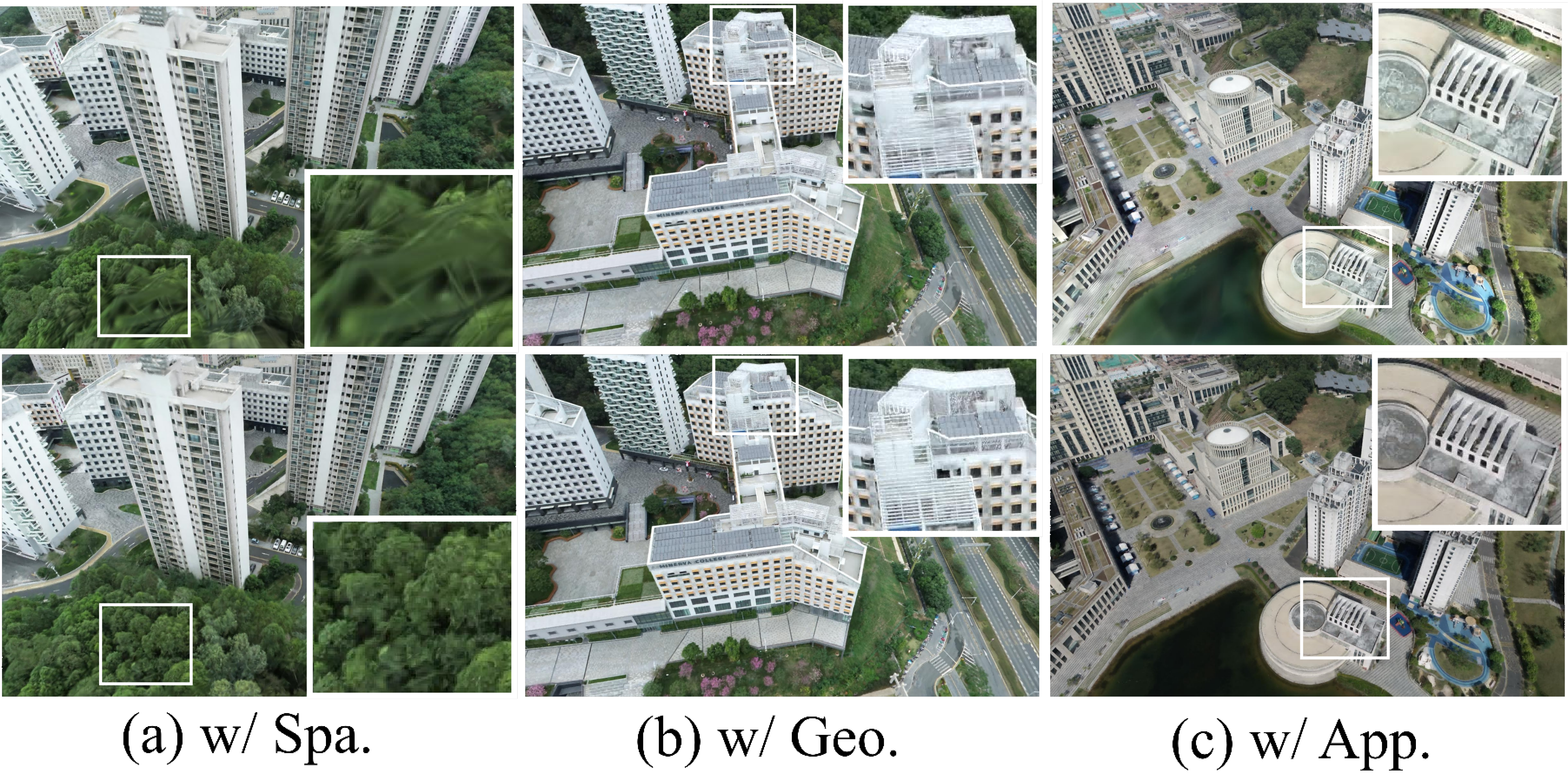} 
\caption{\textbf{Visualization results of ablation study.} The top row shows the results without the corresponding modules, while the bottom row shows the results with the modules. 
Further visualizations are available in the supplementary materials.
}
\label{fig-ab}
\end{figure}

\vspace{-2mm}
\paragraph{Structured Dense Enhancement.} We conduct separate ablation studies on the pointmap-assisted initialization (w/o Ini.) and sparsity compensation (w/o Spa.) within the structured dense enhancement module. Results indicate that removing the former leads to a relatively significant performance degradation, whereas the latter only exhibits a slight drop. The performance change in both cases directly correlates with the reduced number of final reconstructed Gaussians. Fig.~\ref{fig-ab}(a) visually corroborates the efficacy of sparsity compensation, showing clear improvement in sparsely observed regions of the scene.

\vspace{-2mm}
\paragraph{Progressive Hybrid Geometric Refinement.} We ablate the progressive geometric refinement module by removing the whole module (w/o Geo.), its sub-component, the hybrid multi-view refinement (w/o Mul.), and the alignment \& restoration operation (w/o Ali.) within it. Removing the whole module yielded the worst F1-score. Furthermore, removing the multi-view refinement or the alignment also significantly impacts geometric metrics. This confirms the critical role of every component. Notably, Fig.~\ref{fig-ab}(b) further illustrates that geometric quality also impacts appearance.

\vspace{-2mm}
\paragraph{Depth-Guided Appearance Modeling.}
Finally, we evaluate the removal of the entire depth-guided appearance modeling module (w/o App.) and the Tri-Mip component (w/o Tri.) within it. Removing appearance modeling caused a substantial performance drop across all metrics, confirming the importance of decoupling geometry from appearance in scenes with inconsistent visual conditions. Further removing the Tri-Mip feature led to an additional decline, with PSNR dropping even further, highlighting the need for geometric awareness in appearance modeling. In contrast, Fig.~\ref{fig-ab}(c) shows that with our appearance modeling, the rendered image become more realistic and natural.

\begin{table}[t]
  \footnotesize
  \caption{\textbf{Ablation on model components.} The experiments are conducted on Russian Building scene of GauU-Scene~\cite{xiong2024gauu} dataset. We adopt a customized 2DGS~\cite{huang20242d} as our base method.}
  \label{tab:ablation}
  \centering
  \setlength{\tabcolsep}{0.1pt}
  \renewcommand{\arraystretch}{1}
  \resizebox{\linewidth}{!}{
  \begin{tabular}{
  >{\centering\arraybackslash}p{1.55cm}
  |>{\centering\arraybackslash}p{1.1cm}
>{\centering\arraybackslash}p{1.1cm}
|>{\centering\arraybackslash}p{0.9cm}
>{\centering\arraybackslash}p{0.9cm}
>{\centering\arraybackslash}p{0.9cm}
|>{\centering\arraybackslash}p{0.9cm}
>{\centering\arraybackslash}p{0.9cm}}
    \toprule
    \multirow{2}{*}{Model} 
    & \multicolumn{2}{c}{Rendering}
    & \multicolumn{3}{|c}{Geometry}
    & \multicolumn{2}{|c}{GS Statistics} \\
    \cmidrule(lr){2-3} \cmidrule(lr){4-6} \cmidrule(lr){7-8}
     & PSNR$\uparrow$ & SSIM$\uparrow$ 
     & P$\uparrow$ & R$\uparrow$ & F1$\uparrow$ 
     & \#G(M) & T(min) \\
    \midrule
    Base
    & 23.88 & 0.774 
    & 0.539 & 0.509 & 0.523 
    & \textbf{4.55} & 134 \\ 
    Base + Para.
    & 24.35 & 0.798 
    & 0.550 & 0.515 & 0.532 
    & 7.30 & \textbf{68} \\ 

    \midrule
    w/o Ini.
    & 24.84 & 0.808 
    & 0.598 & 0.557 & 0.577 
    & 7.51 & 98 \\
    w/o Spa.
    & 24.88 & 0.811 
    & 0.608 & 0.560 & 0.583 
    & 8.02 & 104 \\

    \midrule
    w/o Geo. 
    & 24.83 & 0.807
    & 0.571 & 0.557 & 0.564
    & 8.99 & 89 \\ 
    w/o Mul.
    & 24.87 & 0.810 
    & 0.586 & 0.556 & 0.571 
    & 8.17 & 87 \\
    w/o Ali. 
    & 24.86 & 0.811 
    & 0.603 & 0.559 & 0.580 
    & 8.18 & 101 \\

    \midrule
    w/o App.
    & 24.46 & 0.807 
    & 0.581 & 0.543 & 0.562 
    & 8.29 & 99 \\
    w/o Tri.
    & 23.96 & 0.807 
    & 0.590 & 0.549 & 0.569 
    & 8.08 & 95 \\

    \midrule
    Full Model 
    & \textbf{24.94} & \textbf{0.814} 
    & \textbf{0.610} & \textbf{0.562} & \textbf{0.585} 
    & 8.20 & 106 \\
    \bottomrule
  \end{tabular}
} 
\end{table}

\section{Conclusion}
In this paper, we present MetroGS, a novel Gaussian Splatting framework designed for large-scale scene reconstruction. Leveraging the foundation of distributed 2DGS, we integrate a structured dense enhancement scheme, a progressive hybrid geometric refinement strategy, and a depth-guided appearance modeling module. Together, these components enable geometrically accurate and training-efficient reconstruction. Extensive experiments on multiple large-scale scene datasets validate the efficacy of our method, demonstrating superior reconstruction performance.

\section*{Acknowledgements}

This work was in part supported by the Chinese Academy of Sciences, the Strategic Priority Research Program of the Chinese Academy of Sciences (XDA0450402), the Beijing Natural Science Foundation (L259015), the National Natural Science Foundation of China (62172392), and the Innovation Research Program of ICT CAS (E261070).

%% file: sec/X_suppl.tex
\clearpage
\setcounter{page}{1}
\maketitlesupplementary

\appendix
\renewcommand{\thesection}{\Alph{section}}

\section{Implementation Details} \label{secA}
For the GauU-Scene~\cite{xiong2024gauu} dataset, we conducted parallel training with a batch size of 4, targeting a total of 60,000 iterations. Subsequently, we train both the single-view and multi-view geometric refinement stages of 
$\mathcal{L}_{geo}$ for 30,000 iterations. During this process, $\lambda_d$ decreases from 0.5 to 0.005 as training progresses, while $\lambda_n$ is set to 0.0125, $\lambda_s$ to 0.1, and $\lambda_{mv}$ to 2.5. For $\mathcal{L}_{app}$, the weight $\lambda$ is set to 0.8. Densification terminates after the 15,000th iteration, with sparsity compensation parameters set to $S_{\text{th}}=20$ and $V_{\text{th}}=10$. The voxel size is set to 0.1 or 0.01 depending on the scale of the scene. For evaluation, only the view embeddings from the training set are available. Since the image filenames encode temporal information, we first use it to identify the two training views that are temporally closest to each test view. We select the candidate with the most similar camera pose to the test view. This nearest-neighbor assignment provides the interpolated view embedding for the test view.

For the MatrixCity~\cite{li2023matrixcity} dataset, the Aerial and Street scenes were trained for 150,000 and 180,000 iterations, respectively. For $\mathcal{L}_{geo}$, single-view optimization is performed until the 50,000th iteration, followed by the switch to multi-view refinement. Densification is also terminated at the 50,000th iteration. All other training configurations follow those used for the GauU-Scene dataset. For evaluation, test image filenames lack temporal information, so each test view selects its most relevant training view solely based on camera-pose similarity. The corresponding view embedding is then used for image rendering.

For geometric quality evaluation, we follow the parameter settings used in CityGSV2~\cite{liu2024citygaussianv2}. Specifically, we render RGB images and depth maps from the training viewpoints and fuse them into a projected truncated signed distance functio (TSDF) volume~\cite{zeng20173dmatch} to extract surface meshes and point clouds. GauU-Scene uses a voxel size of 0.01, an SDF truncation of 0.04, and a depth truncation of 2.0. In MatrixCity, the Aerial split uses 0.01 / 0.04 / 5.0 for voxel size, SDF truncation, and depth truncation, respectively, whereas the Street split adopts 1 / 4 / 500.

\section{Hyperparameters of Other Methods}
For the visualization results of 2DGS, CityGS, and CityGSV2, we train the models using the default parameter settings provided in the CityGSV2 codebase, and for CityGSV2, we use the provided checkpoints. For the comparison with CityGS-$\mathcal{X}$, we utilized its provided Mill19 configuration to train the GauU-Scene dataset. Crucially, we disabled the progressive LOD (Level of Detail) training within this configuration to ensure better preservation of scene details. For the MatrixCity dataset, we directly applied the corresponding official configuration provided by CityGS-$\mathcal{X}$ for training.

\section{Supplementary Method Description}

\begin{figure}[t]
  \centering
  \includegraphics[width=.99\linewidth]{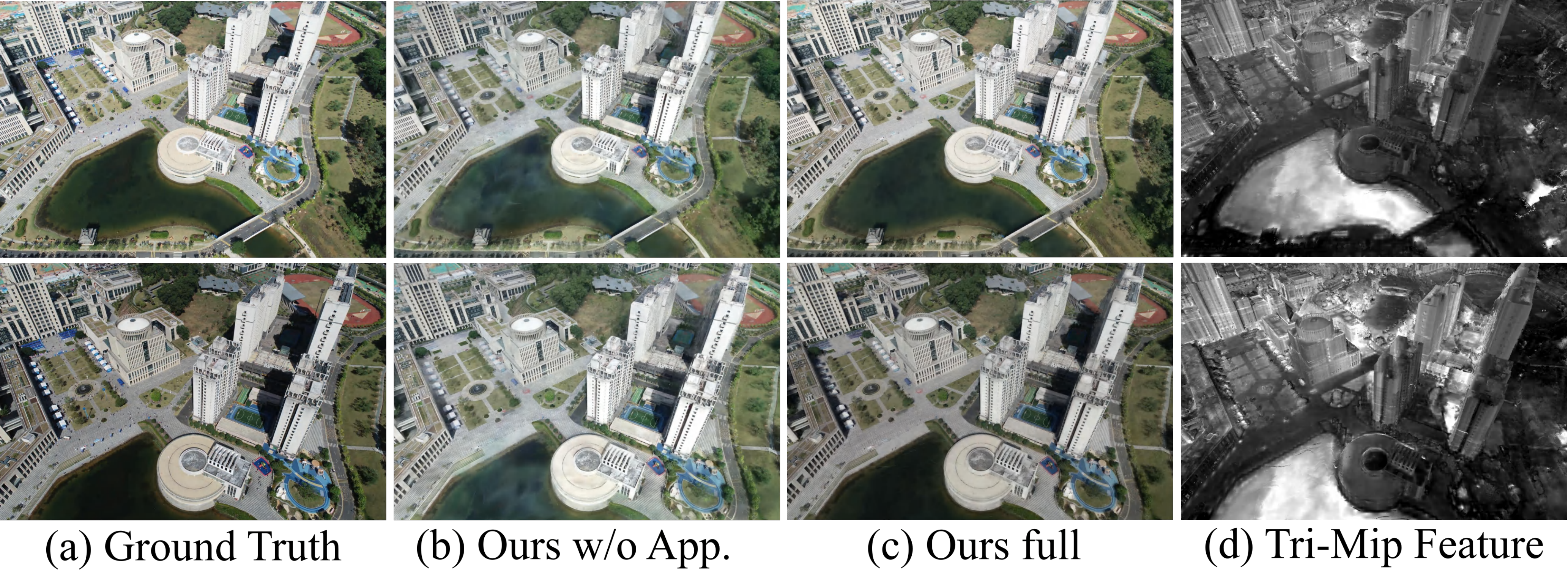}
  \vspace{-4pt}
  \caption{Effectiveness of the appearance modeling.}
  \vspace{-15pt}
  \label{trimipa}
\end{figure}

\subsection{Hybird Multi-View Refinement}

During the multi-view consistency optimization stage, we first estimate an initial refined depth using Multi-View PatchMatch. The algorithm requires neighboring views. For each reference view, we preselect up to eight candidates based on camera orientation and spatial proximity, and randomly sample $K \geq 1$ views during multi-view optimization. Our implementation follows the multi-scale design of ACMM~\cite{xu2019multi}, but keeps the image resolution fixed while progressively reducing the patch size. Since PatchMatch operates on rendered depth, it is sensitive to local errors caused by floating Gaussians. To mitigate this issue, we enlarge the hypothesis sampling range and exclude the central region, adopting a non-local sampling strategy similar to DPE-MVS~\cite{chen2025dual}. This design yields a reliable initial refined depth, even in large weakly textured regions.

Then we compute reprojection errors using neighboring rendered depths to filter out geometrically inconsistent estimates. Although effective, this step may remove correct depths due to occlusion or rendering noise. To recover such regions, we leverage monocular depth via local alignment. Specifically, we partition the image into non-overlapping image blocks and perform local alignment between the monocular depth and the filtered depth within each block (Eq.~\ref{eq8}). Removed depths are restored if consistent with the aligned monocular depth. To improve robustness and completeness, we apply this alignment in a coarse-to-fine recursive manner with progressively smaller blocks. Finally, we obtain the desired depth map $\mathcal{D}_{mv}$, which is used to provide refined supervision for enforcing multi-view consistency.

\subsection{Depth-Guided Appearance Modeling}

Fig.~\ref{trimipa} presents visual comparisons, demonstrating the effectiveness of appearance decoupling. To achieve this, we adopt a Tri-Mip structure to encode geometry-related features, which are queried in 3D space using depth information. The triplane representation ensures that the same spatial location yields consistent geometric features across different views. The mip mechanism further adapts features based on the distance to the current view, enabling consistent and accurate color mapping across varying scales and resolutions. In addition, we introduce view-specific appearance embeddings for each training view to avoid ambiguity in Gaussian geometry under appearance variations. This design incurs low computational overhead, is independent of the number of Gaussians, and can be removed after training.

\begin{table}[t]
\centering
\caption{\textbf{Efficiency performance comparison on the GauU-Scene~\cite{xiong2024gauu} dataset.} Entries marked with an asterisk (*) represent the intermediate results obtained after 30,000 training iterations.}
\label{tab:efficiency}
\resizebox{\linewidth}{!}{
\begin{tabular}{l |l |c c |c c}
    \toprule
    Scene & Method & $\text{PSNR}\uparrow$ & $\text{F1}\uparrow$ & $\#\text{G(M)}$ & $\text{T(min)}$ \\
    \midrule
    \multirow{4}{*}{Russian} 
    & V2-coarse* & 23.46 & 0.509 & \textbf{7.98} & 110 \\
    & Ours* & \textbf{24.60} & \textbf{0.559} & 8.20 & \textbf{50} \\
    \cmidrule(lr){2-6}
    & CityGSV2 & 24.12 & 0.542 & \textbf{7.77} & 363 \\
    & Ours & \textbf{24.94} & \textbf{0.585} & 8.20 & \textbf{106} \\
    \midrule
    \multirow{4}{*}{Residence}
    & V2-coarse* & 22.09 & 0.437 & \textbf{9.29} & 103 \\
    & Ours* & \textbf{23.96} & \textbf{0.470} & 11.33 & \textbf{78} \\
    \cmidrule(lr){2-6}
    & CityGSV2 & 23.55 & 0.466 & \textbf{8.08} & 311 \\
    & Ours & \textbf{24.51} & \textbf{0.494} & 11.33 & \textbf{156} \\
    \midrule
    \multirow{4}{*}{Morden}
    & V2-coarse* & 25.08 & 0.479 & \textbf{7.61} & 98 \\
    & Ours* & \textbf{26.68} & \textbf{0.508} & 9.27 & \textbf{70} \\
    \cmidrule(lr){2-6}
    & CityGSV2 & 25.79 & 0.492 & \textbf{7.89} & 332 \\
    & Ours & \textbf{27.07} & \textbf{0.524} & 9.27 & \textbf{149} \\
    \bottomrule
\end{tabular}
}
\end{table}

\begin{table}[t]
\centering
\caption{\textbf{Efficiency performance comparison on MatrixCity-Aerial~\cite{li2023matrixcity}.} In CityGS-$\mathcal{X}$, which uses an anchor-based Gaussian representation, ``$\times 10$" denotes the Gaussians derived per anchor.}
\label{tab:efficiency2}
\resizebox{\linewidth}{!}{
\begin{tabular}{l |l | c c |c c}
    \toprule
    Scene & 
    Method & $\text{PSNR}\uparrow$ & $\text{F1}\uparrow$ & $\#\text{G(M)}$ & $\text{T(min)}$ \\
    \midrule
    \multirow{2}{*}{MC-Aerial} 
    & 
    CityGS-$\mathcal{X}$ & \textbf{27.53} & 0.582 & 2.48$\times$10 & 716 \\
    & 
    Ours & 27.52 & \textbf{0.677} & 17.09 & \textbf{415} \\
    \bottomrule
\end{tabular}
}
\end{table}

\section{Additional Results}
\subsection{Training Efficiency Analysis}

Using a system with four RTX 3090 GPUs, we conducted a training efficiency comparison between CityGSV2 and CityGS-$\mathcal{X}$ on the GauU-Scene and MatrixCity-Aerial datasets, respectively. As shown in Tab.~\ref{tab:efficiency}, our method consistently outperforms CityGSV2 in both rendering quality and geometric fidelity, while also demonstrating a significant improvement in training efficiency. Notably, even the intermediate results of our model at 30k iterations already surpass the final performance of CityGSV2, while requiring less than 25\% of its training time. Across the GauU-Scene dataset, our final model achieves an average 2.55× training speedup relative to CityGSV2. Tab.~\ref{tab:efficiency2} presents a comparison of training efficiency between CityGS-$\mathcal{X}$ and our method on the MatrixCity-Aerial dataset. Our approach achieves superior geometric fidelity (F1: 0.677 vs. 0.582) with a 1.7× reduction in training time, while maintaining comparable PSNR performance. Overall, these results highlight the remarkable speed and efficiency of our method. It is worth noting that CityGSV2 and CityGS-$\mathcal{X}$ adopt model-size reduction strategies such as trimming~\cite{fan2024trim} and anchor-based Gaussian compression~\cite{ren2024octree}. Enhancing model-size compactness therefore remains a promising direction for further improving the efficiency of our method.

\subsection{Impact of Prior Choices}
As shown in Tab.~\ref{tab:prior} and the ablation studies in the main paper, incorporating priors leads to clear performance improvements, but the overall gains are not dominated by stronger priors. Further analysis reveals that the PointMap model contributes more significantly to the performance, highlighting the importance of high-quality initial points. In contrast, different monocular depth estimation methods have relatively limited impact on the final results.

\begin{table}[t]
\centering
\caption{\textbf{Ablation study on priors on the Russian Building~\cite{xiong2024gauu}.} Our method is robust to different choices of priors, with only limited impact on final performance.}
\label{tab:prior}
\resizebox{\linewidth}{!}{
\begin{tabular}{c | c | c | c | c}
    \toprule
    Metrics & 
    Ours w/o priors & VGGT + DAV2 & VGGT + MoGe-2 & PI3 + MoGe-2 \\
    \midrule
    PSNR / SSIM
    & 24.85 / 0.807 & 24.94 / 0.812 & 24.93 / 0.812 & 24.94 / 0.814 \\
    F1-score 
    & 0.575 & 0.584 & 0.584 & 0.585 \\
    \bottomrule
\end{tabular}
}
\end{table}

\begin{figure}[t]
\centering
\includegraphics[width=\linewidth]{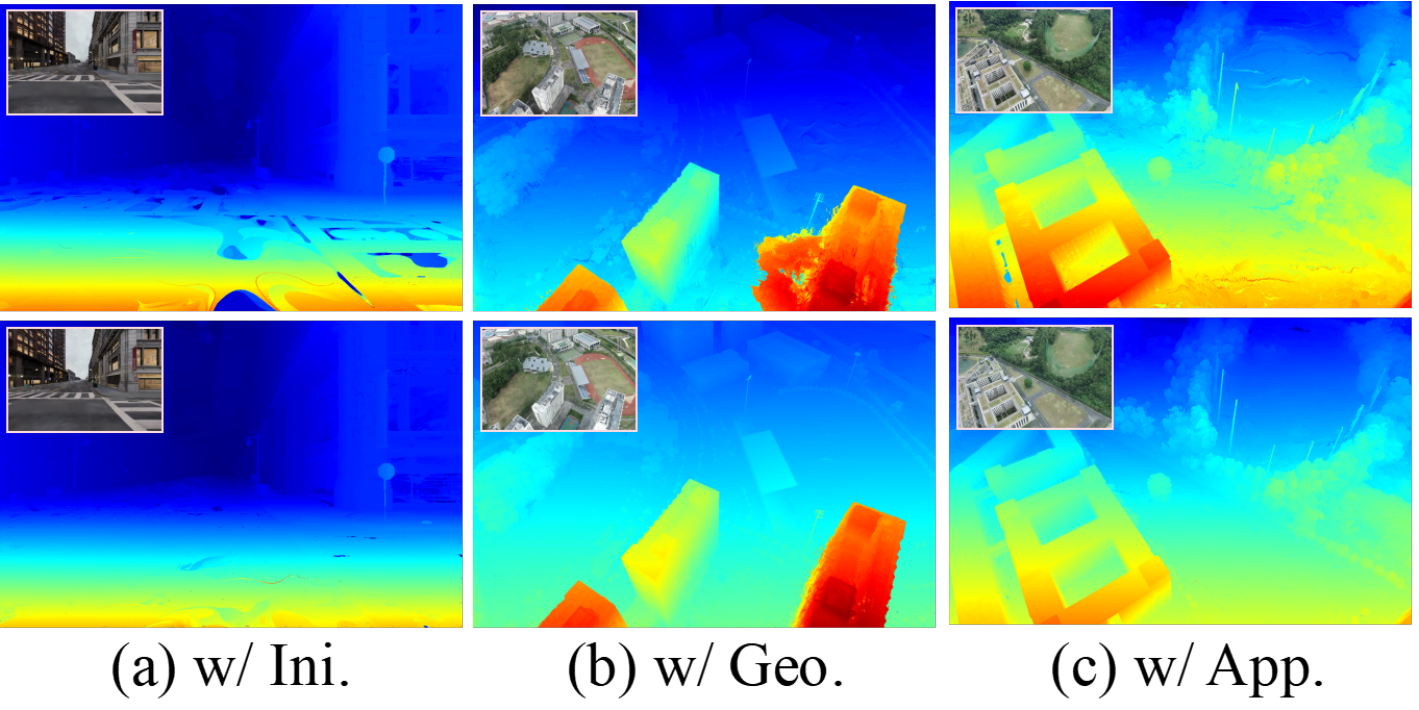} 
\caption{\textbf{Supplementary Visualization of ablation study results.} The top row shows results without the modules, and the bottom row shows results with them. Our components yield a significant improvement in depth quality, effectively addressing challenges across diverse and complex scenes.}
\label{fig-geo}
\end{figure}

\begin{figure*}[t]
\centering
\includegraphics[width=\textwidth]{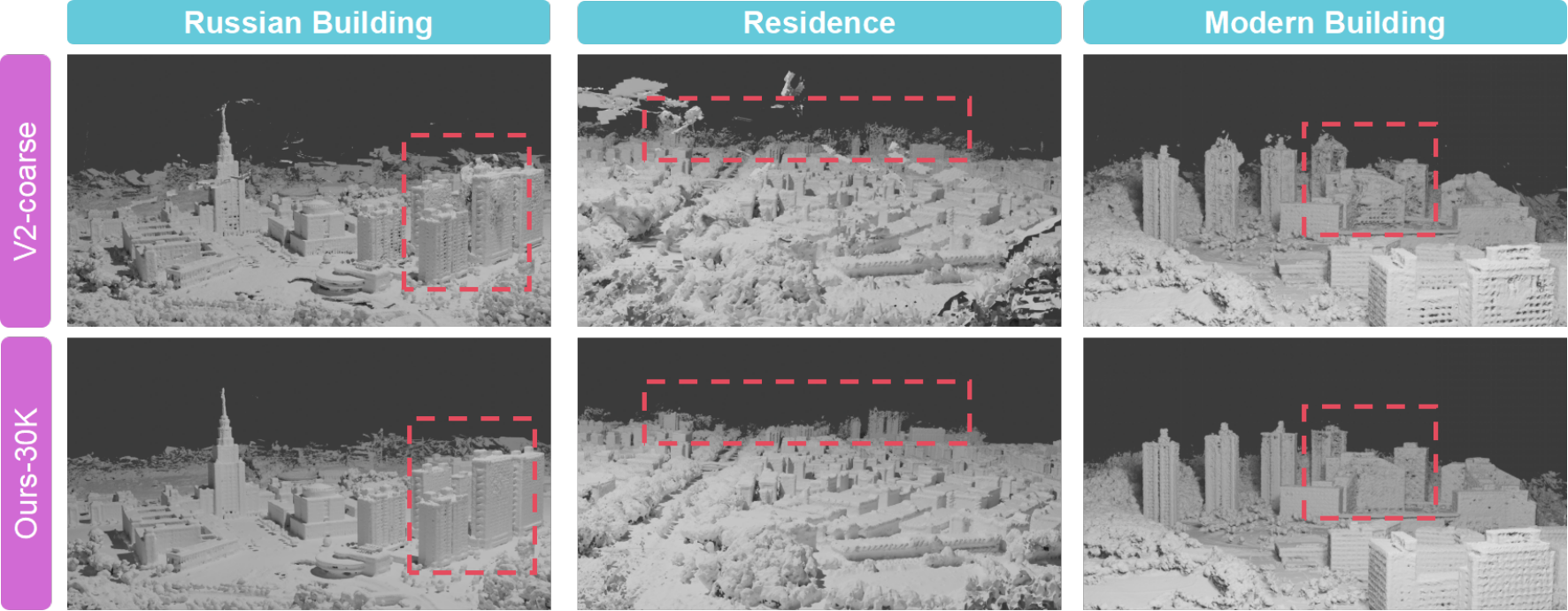} 
\caption{\textbf{Qualitative comparison of meshes on the GauU-Scene~\cite{xiong2024gauu} dataset.} Our method achieves higher-quality results.}
\label{fig-gauu}
\end{figure*}

\begin{figure*}[t]
\centering
\includegraphics[width=\textwidth]{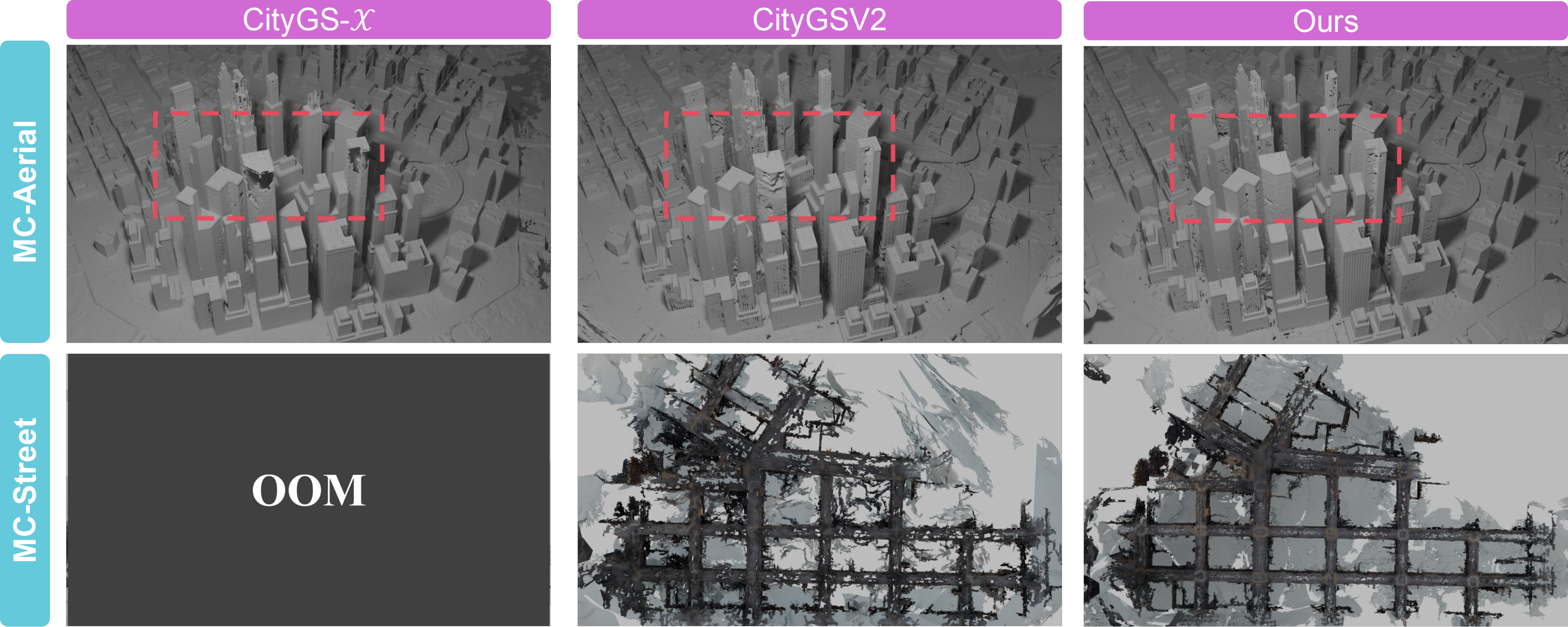} 
\caption{\textbf{Mesh visualization comparison on MatrixCity~\cite{xiong2024gauu}.} Our method provides better results than the baselines.}
\label{fig-mc}
\end{figure*}

\subsection{Additional Qualitative Comparison}
Fig.~\ref{fig-geo} presents visual results for further ablation study. Our adopted pointmap assisted initialization effectively supplements sparse point cloud regions, thereby laying a solid geometric foundation for subsequent reconstruction. Progressive hybrid geometric refinement and depth-guided appearance modeling then collaboratively ensure the final geometric quality exhibits high accuracy and completeness.

In addition, we include more comprehensive qualitative comparisons with the baseline methods. Fig.~\ref{fig-gauu} presents the mesh reconstruction visualization comparison on the GauU-Scene dataset. Given the relatively small size of the image data, we conducted an equivalent comparison in terms of training time: we trained our method for 30,000 iterations and compared its results with those of CityGSV2-coarse. The reconstructed meshes from our method are much cleaner, containing minimal spurious artifacts or floating mesh fragments. Fig~\ref{fig-mc} further presents a comparison of our method's results against CityGSV2 and CityGS-$\mathcal{X}$ on the MatrixCity dataset. The results indicate that our approach achieves a better balance between geometric accuracy and completeness.

\begin{table*}[!t]
  \caption{\textbf{Quantitative results on the Mill19~\cite{turki2022mega} dataset and UrbanScene3D~\cite{lin2022capturing} dataset.} The \textbf{best} and \underline{second best} results are highlighted. All missing results are denoted by a “–”.}
  \label{tab:mill19}
  \centering
  \footnotesize
  \setlength{\tabcolsep}{2pt}
  \renewcommand{\arraystretch}{0.96}
  \resizebox{\textwidth}{!}{
  \begin{tabular}{l|
>{\centering\arraybackslash}p{1cm}
>{\centering\arraybackslash}p{1cm}
>{\centering\arraybackslash}p{1.1cm}
|>{\centering\arraybackslash}p{1cm}
>{\centering\arraybackslash}p{1cm}
>{\centering\arraybackslash}p{1cm}
|>{\centering\arraybackslash}p{1cm}
>{\centering\arraybackslash}p{1cm}
>{\centering\arraybackslash}p{1cm}
|>{\centering\arraybackslash}p{1cm}
>{\centering\arraybackslash}p{1cm}
>{\centering\arraybackslash}p{1cm}}
    \toprule
    \multirow{2}{*}{Methods} 
    & \multicolumn{3}{c}{Building} 
    & \multicolumn{3}{|c}{Rubble} 
    & \multicolumn{3}{|c}{Residence} 
    & \multicolumn{3}{|c}{Sci-Art} \\
    \cmidrule(lr){2-4} \cmidrule(lr){5-7} \cmidrule(lr){8-10} \cmidrule(lr){11-13}
     & PSNR$\uparrow$ & SSIM$\uparrow$ & LPIPS$\downarrow$ 
     & PSNR$\uparrow$ & SSIM$\uparrow$ & LPIPS$\downarrow$
     & PSNR$\uparrow$ & SSIM$\uparrow$ & LPIPS$\downarrow$ 
     & PSNR$\uparrow$ & SSIM$\uparrow$ & LPIPS$\downarrow$ \\
    \midrule
    NeuS
    & 18.01 & 0.463 & 0.611 
    & 20.46 & 0.480 & 0.618
    & 17.85 & 0.503 & 0.533
    & 18.62 & 0.633 & 0.472
    \\
    Neuralangelo
    & 17.89 & 0.582 & 0.322 
    & 20.18 & 0.625 & 0.314
    & 18.03 & 0.644 & 0.263
    & 19.10 & 0.769 & 0.231
    \\
    SuGaR
    & 17.76 & 0.507 & 0.455 
    & 20.69 & 0.577 & 0.453
    & 18.74 & 0.603 & 0.406
    & 18.60 & 0.698 & 0.349
    \\
    PGSR 
    & 16.12 & 0.480 & 0.573 
    & 23.09 & 0.728 & 0.334 
    & 20.57 & 0.746 & 0.289 
    & 19.72 & 0.799 & 0.275
    \\ 
    PGSR+VastGS 
    & 21.63 & 0.720 & 0.300 
    & 25.32 & 0.768 & 0.274 
    & $-$ & $-$ & $-$ 
    & $-$ & $-$ & $-$
    \\
    CityGS
    & 21.55 & 0.778 & 0.246 
    & {25.77} & {0.813} & {0.228}
    & {22.00} & {0.813} & {0.211}
    & {21.39} & {0.837} & {0.230}
    \\
    CityGS-$\mathcal{X}$ 
    & \underline{22.76} & \textbf{0.817} & \underline{0.191}
    & \underline{26.15} & \underline{0.823} & \underline{0.210}
    & \underline{22.44} & \underline{0.819} & \underline{0.194}
    & \underline{22.77} & \underline{0.867} & \underline{0.179}
    \\
    CityGSV2
    & 19.07 & 0.650 & 0.397
    & 23.75 & 0.720 & 0.322
    & 21.15 & 0.769 & 0.234
    & 20.66 & 0.810 & 0.266
    \\ 
    Ours
    & \textbf{23.06} & \underline{0.787} & \textbf{0.173} 
    & \textbf{27.48} & \textbf{0.826} & \textbf{0.147} 
    & \textbf{23.38} & \textbf{0.824} & \textbf{0.166} 
    & \textbf{25.96} & \textbf{0.872} & \textbf{0.152}
    \\
    \bottomrule
  \end{tabular}
} 
\end{table*}

\subsection{Additional Dataset Evaluation}

We have also conducted supplementary evaluations on the Mill19~\cite{turki2022mega} and UrbanScene3D~\cite{lin2022capturing} datasets, which are widely used for assessing rendering quality in the field of large-scale scene reconstruction. Four scenes were selected: Building, Rubble, Residence, and Sci-Art. The configuration uses 100,000 training iterations, with 50,000 iterations allocated to each of the two geometric optimization stages. The densification process is terminated at the 30,000th iteration. The weight $\lambda_s$ set to 0.001. The remaining settings follow those used for GauU-Scene, as detailed in Sec.~\ref{secA}.

\begin{figure}[t]
\centering
\includegraphics[width=\linewidth]{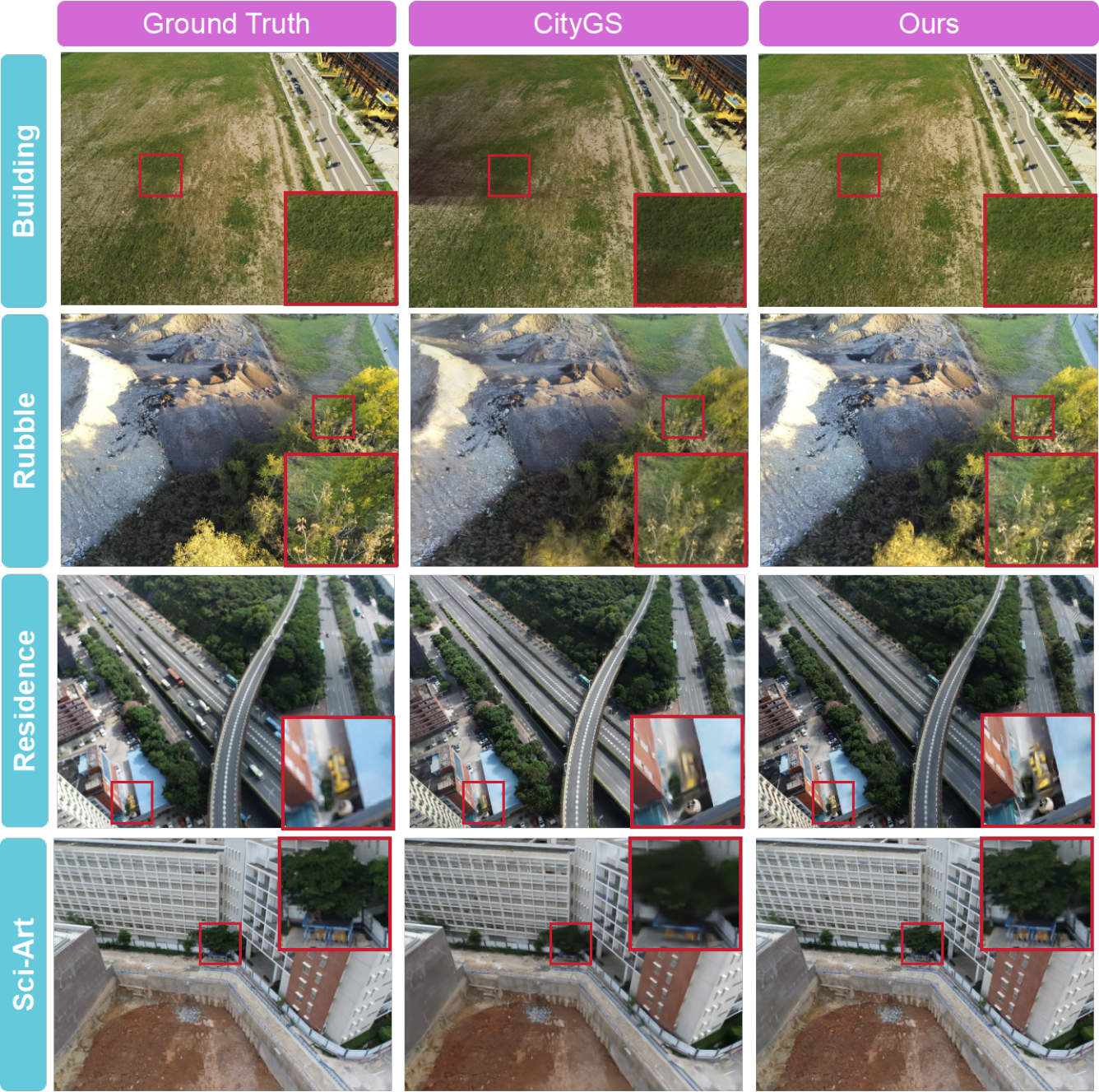} 
\caption{\textbf{Qualitative results on Mill-19~\cite{turki2022mega} and Urbanscene3D~\cite{lin2022capturing}  datasets}. We compare against CityGS.}
\label{fig-mill}
\vspace{-5pt}
\end{figure}

Quantitative results are presented in Tab.~\ref{tab:mill19}, where we compare against other state-of-the-art surface reconstruction methods. Our method achieves state-of-the-art performance among surface reconstruction approaches in terms of PSNR and LPIPS, and ranks first in SSIM for most scenes. In addition, Fig.~\ref{fig-mill} provides a qualitative comparison among our method and CityGS (Public Checkpoints), showing that our approach performs better under challenging illumination conditions and renders fine-grained details more faithfully. Overall, our method achieves superior visual quality and robustness.

\section{Limitation and Discussion}
Although our method achieves efficient training, accurate geometry, and high-quality rendering for large-scale scene reconstruction, it still has several limitations. First, due to hardware constraints, memory consumption remains the primary bottleneck limiting the training scale, which in turn restricts the model’s performance upper bound. Therefore, it is necessary to incorporate techniques such as advanced pruning~\cite{mallick2024taming} and cache management~\cite{zhao2025clm} to alleviate memory pressure. In addition, our method is based on 2DGS. While it demonstrates strong performance in geometric reconstruction, its upper bound in rendering quality may still fall short of 3DGS. To address this, future work could explore introducing new geometric representations similar to~\cite{jiang2025halogs} to achieve full decoupling of geometry and appearance, thereby further improving both geometric accuracy and rendering quality.

Furthermore, as shown in the mesh results of the street-view scene in Fig.~\ref{fig-mc}, although our method achieves better overall performance, some limitations remain. Our results contain fewer floaters, indicating higher geometric accuracy, and benefit from pointmap to produce more complete reconstructions. However, training under street-view settings is more prone to exceeding GPU memory limits. A likely reason is that a large number of occluded Gaussians are still involved in rendering throughout the entire training process, leading to increased memory consumption. To ensure training stability, we introduce scale regularization, which, however, results in a limited number of large-scale Gaussians. This negatively affects the geometric quality of ground regions, leading to insufficient depth continuity and completeness. In addition, we adopt the median depth in 2DGS, which improves robustness but also compromises depth smoothness to some extent. These issues remain important directions for future improvement.
